\documentclass{article}
\usepackage{ijcai21}

\usepackage{times}
\usepackage{soul}
\usepackage{url}

\usepackage[hidelinks]{hyperref}
\usepackage[utf8]{inputenc}
\usepackage[small]{caption}
\usepackage{graphicx}
\usepackage{amsmath}
\usepackage{amsthm}
\usepackage{booktabs}
\usepackage{algorithm}
\usepackage{algorithmic}
\usepackage{graphicx}
\usepackage{amsmath}
\usepackage{amsmath,amssymb,amsfonts}
\usepackage{algorithm}
\usepackage{booktabs}
\usepackage{algorithmic}
\usepackage{bm}
\usepackage{multirow}
\usepackage{amssymb}
\usepackage{color}
\usepackage{subcaption}
\usepackage{multirow}
\usepackage{amssymb}
\usepackage{subcaption}
\usepackage{mathtools}
\usepackage{makecell}
\usepackage{enumitem}
\usepackage{caption}

\title{Hierarchical Self-supervised Augmented Knowledge Distillation}

\author{
Chuanguang Yang$^{1,2}$
\and
Zhulin An$^{1}$\thanks{Corresponding author} \and
Linhang Cai$^{1,2}$\And
Yongjun Xu$^{1}$
\affiliations
$^1$Institute of Computing Technology, Chinese Academy of Sciences, Beijing, China\\
$^2$University of Chinese Academy of Sciences, Beijing, China
\emails
\{yangchuanguang, anzhulin, cailinhang19g, xyj\}@ict.ac.cn
}

\begin{document}

\maketitle

\begin{abstract}
  Knowledge distillation often involves how to define and transfer knowledge from teacher to student effectively. Although recent self-supervised contrastive knowledge achieves the best performance, forcing the network to learn such knowledge may damage the representation learning of the original class recognition task. We therefore adopt an alternative self-supervised augmented task to guide the network to learn the joint distribution of the original recognition task and self-supervised auxiliary task. It is demonstrated as a richer knowledge to improve the representation power without losing the normal classification capability. Moreover, it is incomplete that previous methods only transfer the probabilistic knowledge between the final layers. We propose to append several auxiliary classifiers to hierarchical intermediate feature maps to generate diverse self-supervised knowledge and perform the one-to-one transfer to teach the student network thoroughly. Our method significantly surpasses the previous SOTA SSKD with an average improvement of 2.56\% on CIFAR-100 and an improvement of 0.77\% on ImageNet across widely used network pairs. Codes are available at~\url{https://github.com/winycg/HSAKD}.
\end{abstract}

\section{Introduction}
Orthogonal to efficient network architecture designs~\cite{yang2019multi,zhu2019eena,yang2020gated}, Knowledge Distillation (KD)~\cite{hinton2015distilling} aims to transfer knowledge from a pre-trained high-capacity teacher network to a light-weight student network. The student's performance can often be improved significantly, benefiting from the additional guidance compared with the independent training. The current pattern of KD can be summarized as two critical aspects: (1) what kind of knowledge encapsulated in teacher network can be explored for KD; (2) How to effectively transfer knowledge from teacher to student.

\begin{figure}[t]
	\centering 
	\begin{subfigure}[t]{0.43\textwidth}
		\centering
		\includegraphics[width=\textwidth]{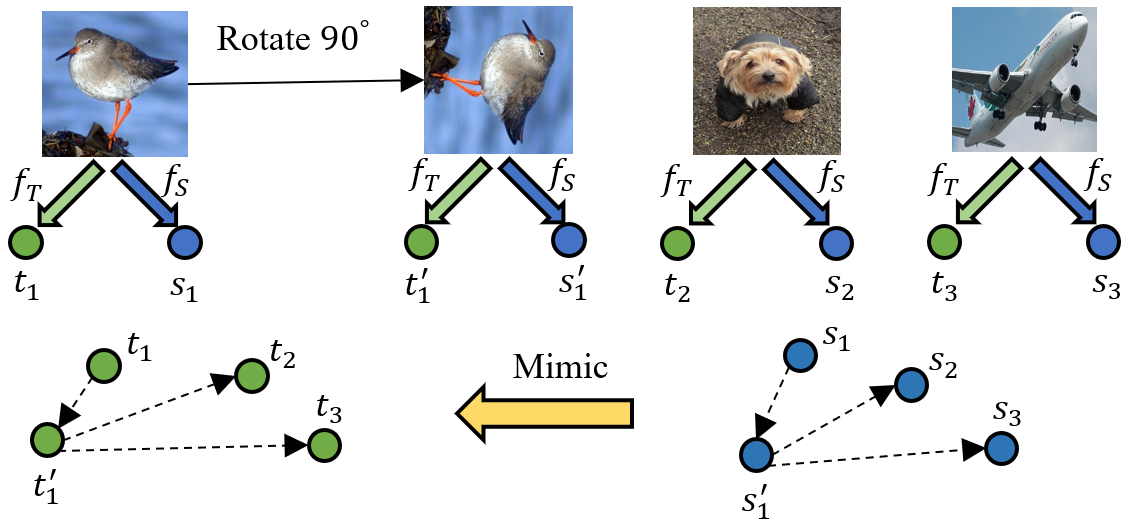}
		\caption{Self-supervised contrastive relationship~\cite{DBLP:conf/eccv/XuLLL20}.}
		\label{scl_}
	\end{subfigure}
	\begin{subfigure}[t]{0.45\textwidth}
		\centering
		\includegraphics[width=\textwidth]{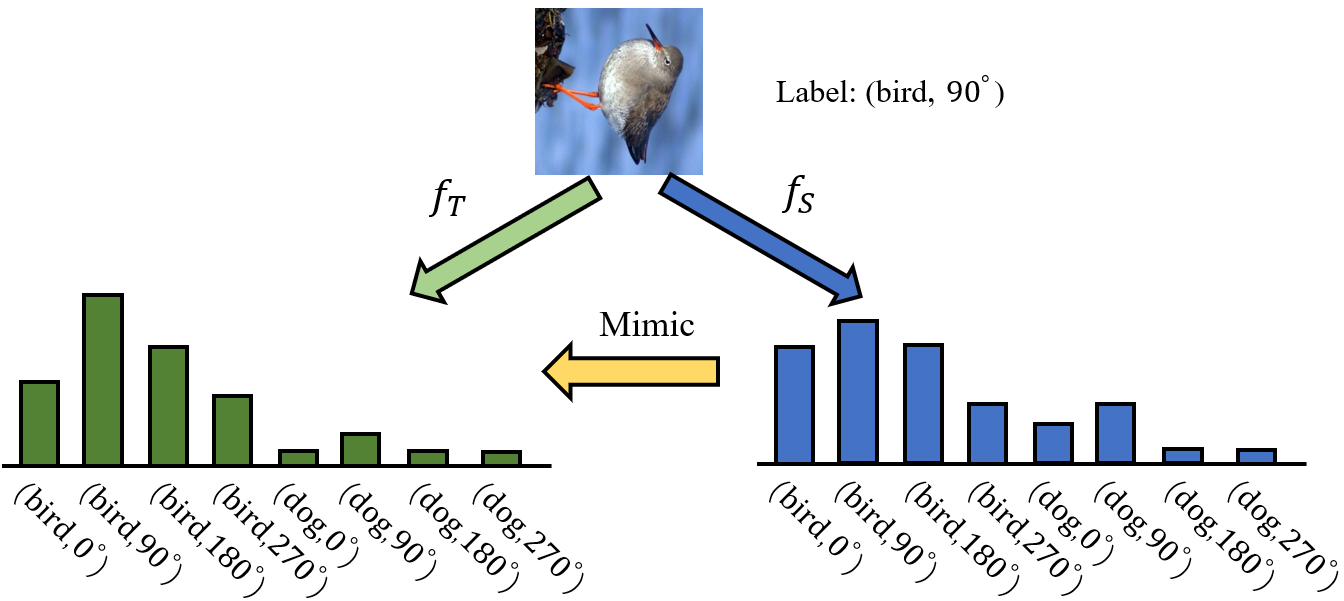}
		\caption{Our introduced self-supervised augmented distribution.}
		\label{icl_}
	\end{subfigure}
	\caption{Difference of self-supervised knowledge between SSKD and our method. (a) SSKD applies contrastive learning by forcing the image and its transformed version closed against other negative images in the feature embedding space. It defines the contrastive relationship as knowledge. (b) Our method unifies the original task and self-supervised auxiliary task into a joint task and defines the self-supervised augmented distribution as knowledge.}
\end{figure} 

The original KD~\cite{hinton2015distilling} minimizes the KL-divergence of predictive class probability distributions between student and teacher networks, which makes intuitive sense to force the student to mimic how a superior teacher generates the final predictions. However, such a highly abstract dark knowledge ignores much comprehensive information encoded in hidden layers. Later works naturally proposed to transfer feature maps~\cite{romero2014fitnets} or their refined information~\cite{zagoruyko2016paying,heo2019knowledge,Sungsoo19Variational} between intermediate layers of teacher and student. A reasonable interpretation of the success of feature-based distillation lies in that hierarchical feature maps throughout the CNN represent the intermediate learning process with an inductive bias of the final solution. Beyond knowledge alignment limited in the individual sample, more recent works~\cite{peng2019correlation,tian2019contrastive} leverage cross-sample correlations or dependencies in high-layer feature embedding space. Inspired by the recent success of self-supervised visual representational learning~\cite{chen2020simple}, SSKD~\cite{DBLP:conf/eccv/XuLLL20} introduces an auxiliary self-supervised task to extract richer knowledge. As shown in Fig.~\ref{scl_}, SSKD proposes transferring cross-sample self-supervised contrastive relationships, making it achieve superior performance in the field of KD.

\begin{table}
	\centering
	\resizebox{0.85\linewidth}{!}{
		\begin{tabular}{cccc}
			\toprule
			Dataset  & Baseline &+DA (Rotation) &+SAL (Rotation)\\
			\midrule
			CIFAR-100  & 78.01 &77.75$_{(\downarrow -0.26)}$&\textbf{79.76}$_{(\uparrow +1.75)}$  \\
			TinyImageNet  & 63.69 &62.66$_{(\downarrow -1.03)}$ &\textbf{65.81}$_{(\uparrow +2.12)}$  \\
			\bottomrule
	\end{tabular}}
	\caption{Top-1 accuracy on ResNet-18 using rotation as a data augmentation (DA) and a self-supervised augmented label (SAL).}
	\label{compar_ss}
\end{table}

However, forcing the network to learn invariant feature representations among transformed images using a self-supervised pretext task with random rotations from $0^{\circ},90^{\circ},180^{\circ},270^{\circ}$ utilized in SSKD may destroy the original visual semantics (\emph{e.g.} 6 v.s. 9). It would increase the difficulty of representation learning for semantic recognition tasks. As validated in Table~\ref{compar_ss}, applying random rotation as an additional data augmentation degrades the classification performance, especially on more challenging TinyImageNet. 

To effectively learn knowledge from self-supervised representation learning without interfering with the original fully-supervised classification task, we use a unified task by combining the label spaces of the original task and self-supervised task into a joint label space, as shown in Fig.~\ref{icl_}. This task is partly inspired by the previous seminal self-supervised representation learning~\cite{Spyros2018Unsupervised,lee2020self}. We further introduce these prior works to explore more powerful knowledge for distillation. To verify the effectiveness of the self-supervised augmented label, we also conduct initial exploratory experiments on standard image classification in Table~\ref{compar_ss}. We find that the performance can be significantly improved by SAL, which can be attributed to learned better feature representations from an extra well-combined self-supervised task.  The good performance further motivates us to define the self-supervised augmented distribution as a promising knowledge for KD.

Another valuable problem lies in how to transfer the probabilistic knowledge between teacher and student effectively. Vanilla KD aligns probability distributions only in the final layer but ignores comprehensive knowledge. Feature-based distillation methods provide one-to-one matching between the same convolutional stages of teacher and student. However, matched feature maps may have different semantic abstractions and result in a negative supervisory effect~\cite{passalis2020heterogeneous}. Compared with feature information, the probability distribution is indeed a more robust knowledge for KD, especially when existing a large architecture gap between teacher and student~\cite{tian2019contrastive}. 

However, it is difficult to explicitly derive comprehensive probability distributions from hidden layers over the original architecture. Therefore a natural idea is to append several auxiliary classifiers to the network at various hidden layers to generate multi-level probability distributions from hierarchical feature maps. It allows us to perform comprehensive one-to-one matching in hidden layers in terms of probabilistic knowledge. Moreover, it is also noteworthy that the gap of abstraction level of any matched distributions would be easily reduced due to the delicately-designed auxiliary classifiers.

We guide all auxiliary classifiers attached to the original network to learn informative self-supervised augmented distributions. Furthermore, we perform Hierarchical Self-supervised Augmented Knowledge Distillation (HSAKD) between teacher and student towards all auxiliary classifiers in a one-to-one manner. By taking full advantage of richer self-supervised augmented knowledge, the student can be guided to learn better feature representations. Note that all auxiliary classifiers are only used to assist knowledge transfer and dropped during the inference period. The overall contributions are summarized as follows:
\begin{itemize}[noitemsep,nolistsep,,topsep=0pt,parsep=0pt,partopsep=0pt]
	\item We introduce a self-supervised augmented distribution that encapsulates the unified knowledge of the original classification task and auxiliary self-supervised task as the richer dark knowledge for the field of KD. 
	\item We propose a one-to-one probabilistic knowledge distillation framework by leveraging the architectural auxiliary classifiers, facilitating comprehensive knowledge transfer and alleviating the mismatch problem of abstraction levels when existing a large architecture gap.
	\item HSAKD significantly refreshes the results achieved by previous SOTA SSKD on standard image classification benchmarks. It can also learn well-general feature representations for downstream semantic recognition tasks. 
\end{itemize}
\begin{figure*}[tbp]  
	\centering  
	\includegraphics[width=1\linewidth]{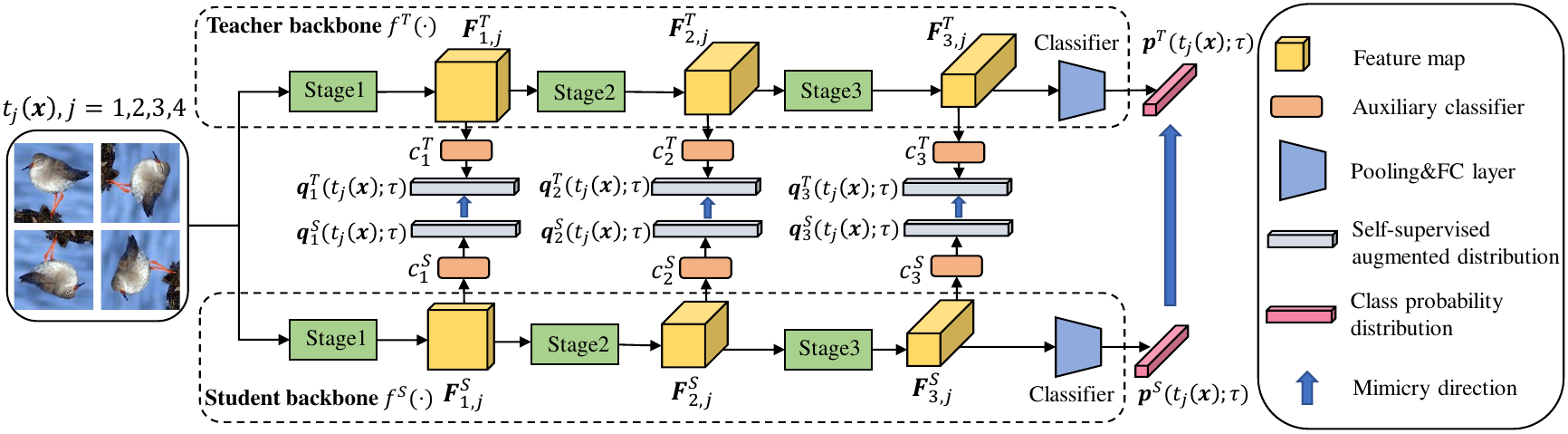}
	\caption{Overview of our proposed HSAKD. Both teacher and student networks are equipped with several auxiliary classifiers after various convolutional stages to capture diverse self-supervised augmented knowledge from hierarchical feature maps. Mimicry loss is applied from self-supervised augmented distributions of the student $\{\bm{q}_{l}^{S}(t_{j}(\bm{x});\tau)\}_{l=1}^{L}$ to corresponding that of the teacher $\{\bm{q}_{l}^{T}(t_{j}(\bm{x});\tau)\}_{l=1}^{L}$ generated from same feature hierarchies in a one-to-one manner. Following the conventional KD, we also force the mimicry from the class probability distribution of student $\bm{p}^{S}(t_{j}(\bm{x});\tau)$ to that of the teacher $\bm{p}^{T}(t_{j}(\bm{x});\tau)$. During the inference period, we only retain the student backbone $f^{S}(\cdot)$ and drop all auxiliary classifiers $\{c^{S}(\cdot)\}_{l=1}^{L}$. Therefore there has no extra inference cost compared with the original student network.}  
	\label{HSKD}

\end{figure*}
\section{Related Work}
\paragraph{Knowledge Distillation.} The seminal KD~\cite{hinton2015distilling} popularizes the pattern of knowledge transfer with a soft probability distribution. Later methods further explore feature-based information encapsulated in hidden layers for KD, such as intermediate feature maps~\cite{romero2014fitnets}, attention maps~\cite{zagoruyko2016paying}, gram matrix~\cite{Gift17} and activation boundaries~\cite{heo2019knowledge}. More recent works explore cross-sample relationship using high-level feature embeddings with various definitions of edge weight~\cite{park2019relational,peng2019correlation,Tung2019Similarity}. The latest SSKD~\cite{DBLP:conf/eccv/XuLLL20} extracts structured knowledge from the self-supervised auxiliary task. Beyond knowledge exploration, \cite{Sungsoo19Variational,tian2019contrastive} maximize the mutual information between matched features. \cite{yang2021multi} further extend this idea into online KD in a contrastive-based manner. To bridge the gap between the teacher and student, \cite{passalis2020heterogeneous} introduce teacher assistant models for smoother KD. However, extra teacher models increase the complexity of the training pipeline. Therefore, we choose to append several well-designed auxiliary classifiers to alleviate the knowledge gap and facilitate comprehensive knowledge transfer.
\paragraph{Self-supervised Representational Learning (SRL).} The seminal SRL popularizes the pattern that guides the network to learn which transformation is applied to a transformed image for learning feature representations. Typical transformations can be rotations~\cite{Spyros2018Unsupervised}, jigsaw~\cite{noroozi2016unsupervised} and colorization~\cite{zhang2016colorful}. More recently, \cite{misra2020self,chen2020simple} learn invariant feature representations under self-supervised pretext tasks by maximizing the consistency of representations among various transformed versions of the same image. Both SSKD and our HSAKD are related to SRL. SSKD uses the latter SRL pattern to extract knowledge. In contrast, HSAKD combines the former classification-based pattern of SRL with the fully-supervised classification task to extract richer joint knowledge.

\section{Method}
\subsection{Self-supervised Augmented Distribution}
We present the difference between the original class probability distribution and self-supervised augmented distribution using a conventional classification network of CNN. A CNN can be decomposed into a feature extractor $\Phi(\cdot;\bm{\mu})$ and a linear classifier $g(\cdot;\bm{w})$, where $\bm{\mu}$ and $\bm{w}$ are weight tensors. Given an input sample $\bm{x}\in \mathcal{X}$, $\mathcal{X}$ is the training set, $\bm{z}=\Phi(\bm{x};\bm{\mu})\in \mathbb{R}^{d}$ is the extracted feature embedding vector, where $d$ is the embedding size. We consider a conventional $N$-way object classification task with the label space $\mathcal{N}=\{1,\cdots,N\}$. The linear classifier attached with softmax normalization maps the feature embedding $\bm{z}$ to a predictive class probability distribution $\bm{p}(\bm{x};\tau)=\sigma(g(\bm{z};\bm{w})/\tau)\in \mathbb{R}^{N}$ over the label space, where $\sigma$ is the softmax function, weight matrix $\bm{w}\in \mathbb{R}^{N\times d}$, $\tau$ is a temperature hyper-parameter to scale the smoothness of distribution.

We introduce an additional self-supervised task to augment the conventional supervised object class space. Learning such a joint distribution can force the network to generate more informative and meaningful predictions benefiting from the original and auxiliary self-supervised task simultaneously. Assuming that we define $M$ various image transformations $\{t_{j}\}_{j=1}^{M}$ with the label space $\mathcal{M}=\{1,\cdots,M\}$, where $t_{1}$ denotes the identity transformation, \emph{i.e.} $t_{1}(\bm{x})=\bm{x}$. To effectively learn composite knowledge, we combine the class space from the original supervised object recognition task and self-supervised task into a unified task. The label space of this task is $\mathcal{K}=\mathcal{N}\otimes \mathcal{M}$, here $\otimes$ is the Cartesian product. $\left | \mathcal{K} \right |=N*M$, where $\left | \cdot \right |$ is the cardinality of the label collection, * denotes element-wise multiplication.

Given a transformed sample $\tilde{\bm{x}}\in \{t_{j}(\bm{x})\}_{j=1}^{M}$ by applying one transformation on $\bm{x}$, $\tilde{\bm{z}}=\Phi(\tilde{\bm{x}};\bm{\mu})\in \mathbb{R}^{d}$ is the extracted feature embedding vector, $\bm{q}(\tilde{\bm{x}};\tau)=\sigma(g(\tilde{\bm{z}};\bm{w})/\tau)\in \mathbb{R}^{N*M}$ is the predictive distribution over the joint label space $\mathcal{K}$, where weight tensor $\bm{w}\in \mathbb{R}^{(N*M)\times d}$. We use $\bm{p}\in \mathbb{R}^{N}$ to denote the normal class probability distribution and $\bm{q}\in \mathbb{R}^{N*M}$ to denote the self-supervised augmented distribution.

\subsection{Auxiliary Architecture Design}
It has been widely known that feature maps with various resolutions encode various patterns of representational information. Higher-resolution feature maps often present more fine-grained object details, while lower-resolution ones often contain richer global semantic information. To take full advantage of hierarchical feature maps encapsulated in a single network, we append several intermediate auxiliary classifiers into hidden layers to learn and distill hierarchical self-supervised augmented knowledge.

For ease of notation, we denote a conventional classification network as $f(\cdot)$, which maps a input sample $t_{j}(\bm{x}),j\in \mathcal{M}$ to the vanilla class probability distribution $\bm{p}(t_{j}(\bm{x});\tau)=\sigma(f(t_{j}(\bm{x}))/\tau)\in \mathbb{R}^{N}$ over the original class space $\mathcal{N}$. Modern CNNs typically utilize stage-wise convolutional blocks to gradually extract coarser features as the depth of the network increases. For example, popular ResNet-50 for ImageNet classification contains consecutive four stages, and extracted feature maps produced from various stages have different granularities and patterns. Assuming that a network contains $L$ stages, we choose to append an auxiliary classifier after each stage, thus resulting in $L$ classifiers $\{c_{l}(\cdot)\}_{l=1}^{L}$, where $c_{l}(\cdot)$ is the auxiliary classifier after $l$-th stage. $c_{l}(\cdot)$ is composed of stage-wise convolutional blocks, a global average pooling layer and a fully connected layer. Denoting the extracted feature map after $l$-th stage as $\bm{F}_{l}$, we can obtain the self-supervised augmented distribution inferred by $c_{l}(\cdot)$ as $\bm{q}_{l}(t_{j}(\bm{x});\tau)=\sigma(c_{l}(\bm{F}_{l}))/\tau)\in \mathbb{R}^{N*M}$ over the joint class space $\mathcal{K}$. The overall design of auxiliary classifiers over a 3-stage network is illustrated in Fig.~\ref{HSKD} for example. \textbf{\emph{The detailed design formulation of various auxiliary classifiers for a specific network can be found in our released codes}}.

\subsection{Training the Teacher Network}
We denote the teacher backbone network as $f^{T}(\cdot)$ and $L$ auxiliary classifiers as $\{c_{l}^{T}(\cdot)\}_{l=1}^{L}$. We conduct an end-to-end training process for preparing the teacher network. On the one hand, we train the $f^{T}(\cdot)$ with normal data $\bm{x}$ by the conventional Cross-Entropy (CE) loss to fit the ground-truth label $y\in \mathcal{N}$, and $\bm{p}^{T}(\bm{x};\tau)=\sigma(f^{T}(\bm{x})/\tau)\in \mathbb{R}^{N}$ is the predictive class probaility distribution. On the other hand, we aim to train $L$ auxiliary classifiers $\{c_{l}^{T}(\cdot)\}_{l=1}^{L}$ for learning hierarchical self-supervised augmented distributions. Given an input sample $t_{j}(\bm{x})$, we feed the feature maps $\{\bm{F}_{l,j}^{T}\}_{l=1}^{L}$ generated from backbone $f^{T}(\cdot)$ to $\{c_{l}^{T}(\cdot)\}_{l=1}^{L}$, respectively. The predictive self-supervised augmented distribution inferred by the $l$-th classifier $c^{T}_{l}$ is $\bm{q}^{T}_{l}(t_{j}(\bm{x});\tau)=\sigma(c^{T}_{l}(\bm{F}_{l,j}^{T}))/\tau)\in \mathbb{R}^{N*M}$. We train all auxiliary classifiers using  CE loss with self-supervised augmented labels across $\{t_{j}(\bm{x})\}_{j=1}^{M}$ as Eq.~(\ref{train_tsad}).
\begin{equation}
\mathcal{L}_{ce\_SAD}^{T}=\frac{1}{M}\sum_{j=1}^{M}\sum_{l=1}^{L}\mathcal{L}_{ce}(\bm{q}^{T}_{l}(t_{j}(\bm{x});\tau),k_{j})
\label{train_tsad}
\end{equation}
Where  $\tau=1$ and $\mathcal{L}_{ce}$ denotes the Cross-Entropy loss. For a bit abuse of notation, we use $k_{j}$ to denote the self-supervised augmented label of $t_{j}(\bm{x})$ in joint class space $\mathcal{K}$. The overall loss for training a teacher is shown in Eq.~(\ref{train_t}).
\begin{equation}
	\mathcal{L}_{T}=\mathbb{E}_{\bm{x}\in \mathcal{X}}[\mathcal{L}_{ce}(\bm{p}^{T}(\bm{x};\tau),y)+\mathcal{L}_{ce\_SAD}^{T}]
	\label{train_t}
\end{equation}

\begin{table*}[t]
	\centering
	\resizebox{1\linewidth}{!}{
		\begin{tabular}{cccccccccc}
			\toprule
			Teacher  & WRN-40-2 &WRN-40-2 &ResNet56 &ResNet32$\times$4 & VGG13 &ResNet50  &WRN-40-2&ResNet32$\times$4   \\
			
			Student  & WRN-16-2 &WRN-40-1 &ResNet20 &ResNet8$\times$4 & MobileNetV2 &MobileNetV2  &ShuffleNetV1&ShuffleNetV2\\
			\midrule
			Teacher  & 76.45 &76.45 &73.44 &79.63 &74.64&79.34&76.45& 79.63\\
			Teacher*  & 80.70 &80.70 &77.20 &83.73 &78.48&83.85&80.70& 83.73\\
			
			Student  & 73.57$_{(\pm 0.23)}$ &71.95$_{(\pm 0.59)}$&69.62$_{(\pm 0.26)}$ &72.95$_{(\pm 0.24)}$ &73.51$_{(\pm 0.26)}$&73.51$_{(\pm 0.26)}$&71.74$_{(\pm 0.35)}$&72.96$_{(\pm 0.33)}$\\
			\midrule
			KD       & 75.23$_{(\pm 0.23)}$ &73.90$_{(\pm 0.44)}$ &70.91$_{(\pm 0.10)}$ & 73.54$_{(\pm 0.26)}$& 75.21$_{(\pm 0.24)}$&75.80$_{(\pm 0.46)}$& 75.83$_{(\pm 0.18)}$&75.43$_{(\pm 0.33)}$    \\
			FitNet& 75.30$_{(\pm 0.42)}$ & 74.30$_{(\pm 0.42)}$  &71.21$_{(\pm 0.16)}$ & 75.37$_{(\pm 0.12)}$&75.42$_{(\pm 0.34)}$&75.41$_{(\pm 0.07)}$& 76.27$_{(\pm 0.18)}$ &76.91$_{(\pm 0.06)}$  \\
			AT    & 75.64$_{(\pm 0.31)}$  & 74.32$_{(\pm 0.23)}$  & 71.35$_{(\pm 0.09)}$& 75.06$_{(\pm 0.19)}$&74.08$_{(\pm 0.21)}$&76.57$_{(\pm 0.20)}$& 76.51$_{(\pm 0.44)}$&76.32$_{(\pm 0.12)}$   \\
			AB& 71.26$_{(\pm 1.32)}$& 74.55$_{(\pm 0.46)}$  &71.56$_{(\pm 0.19)}$ & 74.31$_{(\pm 0.09)}$&74.98$_{(\pm 0.44)}$&75.87$_{(\pm 0.39)}$& 76.43$_{(\pm 0.09)}$&76.40$_{(\pm 0.29)}$\\
			VID	& 75.31$_{(\pm 0.22)}$ & 74.23$_{(\pm 0.28)}$ &71.35$_{(\pm 0.09)}$ &75.07$_{(\pm 0.35)}$ &75.67$_{(\pm 0.13)}$&75.97$_{(\pm 0.08)}$&76.24$_{(\pm 0.44)}$ &  75.98$_{(\pm 0.41)}$  \\
			RKD & 75.33$_{(\pm 0.14)}$ & 73.90$_{(\pm 0.26)}$  &71.67$_{(\pm 0.08)}$ &74.17$_{(\pm 0.22)}$ &75.54$_{(\pm 0.36)}$&76.20$_{(\pm 0.06)}$&75.74$_{(\pm 0.32)}$ & 75.42$_{(\pm 0.25)}$  \\
			SP& 74.35$_{(\pm 0.59)}$ & 72.91$_{(\pm 0.24)}$ &71.45$_{(\pm 0.38)}$ & 75.44$_{(\pm 0.11)}$&75.68$_{(\pm 0.35)}$&76.35$_{(\pm 0.14)}$&76.40$_{(\pm 0.37)}$&76.43$_{(\pm 0.21)}$\\
			CC& 75.30$_{(\pm 0.03)}$ & 74.46$_{(\pm 0.05)}$ &71.44$_{(\pm 0.10)}$ & 74.40$_{(\pm 0.24)}$&75.66$_{(\pm 0.33)}$&76.05$_{(\pm 0.25)}$&     75.63$_{(\pm 0.30)}$&75.74$_{(\pm 0.18)}$\\
			CRD& 75.81$_{(\pm 0.33)}$ & 74.76$_{(\pm 0.25)}$ &\underline{71.83}$_{(\pm 0.42)}$ & 75.77$_{(\pm 0.24)}$&76.13$_{(\pm 0.16)}$&76.89$_{(\pm 0.27)}$&     76.37$_{(\pm 0.23)}$&76.51$_{(\pm 0.09)}$\\
			SSKD& \underline{76.16}$_{(\pm 0.17)}$ & \underline{75.84}$_{(\pm 0.04)}$ & 70.80$_{(\pm 0.02)}$&\underline{75.83}$_{(\pm 0.29)}$ &\underline{76.21}$_{(\pm 0.16)}$&\underline{78.21}$_{(\pm 0.16)}$&\underline{76.71}$_{(\pm 0.31)}$&\underline{77.64}$_{(\pm 0.24)}$\\
			\midrule
			Ours& 77.20$_{(\pm 0.17)}$  & 77.00$_{(\pm 0.21)}$ &72.58$_{(\pm 0.33)}$ & 77.26$_{(\pm 0.14)}$&77.45$_{(\pm 0.21)}$&78.79$_{(\pm 0.11)}$&  78.51$_{(\pm 0.20)}$&79.93$_{(\pm 0.11)}$\\
			Ours*& \textbf{78.67}$_{(\pm 0.20)}$  & \textbf{78.12}$_{(\pm 0.25)}$ &\textbf{73.73}$_{(\pm 0.10)}$ & \textbf{77.69}$_{(\pm 0.05)}$&\textbf{79.27}$_{(\pm 0.12)}$&\textbf{79.43}$_{(\pm 0.24)}$&    \textbf{80.11}$_{(\pm 0.32)}$&\textbf{80.86}$_{(\pm 0.15)}$\\
			\bottomrule\end{tabular}}
	\caption{Top-1 accuracy (\%) comparison of SOTA distillation methods across various teacher-student pairs on CIFAR-100. All results are reproduced by ours using author-provided code. The numbers in \textbf{Bold} and \underline{underline} denote the best and the second-best results, respectively. 'Teacher' denotes that we first train the backbone $f^{T}(\cdot)$ and then train auxiliary classifiers $\{c_{l}^{T}(\cdot)\}_{l=1}^{L}$ based on the frozen $f^{T}(\cdot)$. For a fair comparison, all compared methods and 'Ours' are supervised by 'Teacher'. 'Teacher*' denotes that we train $f^{T}(\cdot)$ and $\{c_{l}^{T}(\cdot)\}_{l=1}^{L}$ jointly, leading to a more powerful teacher network. 'Ours*' denotes the results supervised by 'Teacher*' for pursuing better performance.} 
	\label{Comparisonx}
\end{table*}
Note that the two losses in Eq.~\ref{train_t} have different roles. The first loss aims to simply fit the normal data for learning general classification capability. The second loss aims to generate additional self-supervised augmented knowledge by the existing hierarchical features derived from the backbone network. This method facilitates richer knowledge distillation benefiting from the self-supervised task beyond the conventional fully-supervised task. 
\subsection{Training the Student Network}
We denote the student backbone network as $f^{S}(\cdot)$ and $L$ auxiliary classifiers as $\{c_{l}^{S}(\cdot)\}_{l=1}^{L}$. We conduct an end-to-end training process under the supervision of the teacher network. The overall loss includes the task loss from pre-defined ground-truth labels and the mimicry loss from the pre-trained teacher network. 

\paragraph{Task loss.} We force the $f^{S}(\cdot)$ to fit the normal data $\bm{x}$ as the task loss:
\begin{equation}
\mathcal{L}_{task}=\mathcal{L}_{ce}(\bm{p}^{S}(\bm{x};\tau),y)
\label{train_s}
\end{equation}
 Where $\bm{p}^{S}(\bm{x};\tau)=\sigma(f^{S}(\bm{x})/\tau)\in \mathbb{R}^{N}$ is the predictive class probaility distribution.
We also had tried to force $L$ auxiliary classifiers $\{c_{l}^{S}(\cdot)\}_{l=1}^{L}$ to learn the self-supervised augmented distributions from the joint hard label of the original and self-supervised tasks by $\mathcal{L}_{ce\_SAD}^{S}$ as the additional loss:
\begin{equation}
\mathcal{L}_{ce\_SAD}^{S}=\frac{1}{M}\sum_{j=1}^{M}\sum_{l=1}^{L}\mathcal{L}_{ce}(\bm{q}^{S}_{l}(t_{j}(\bm{x});\tau),k_{j})
\label{train_ss}
\end{equation}
Where $\bm{q}^{S}_{l}(t_{j}(\bm{x});\tau)=\sigma(c^{S}_{l}(\bm{F}_{l,j}^{S}))/\tau)\in \mathbb{R}^{N*M}$, and $\bm{F}_{l,j}^{S}$ is the extracted feature map from the $l$-th stage of $f^{S}(\cdot)$ for the input $t_{j}(\bm{x})$. However, we empirically found that introducing loss~(\ref{train_ss}) into the original task loss~(\ref{train_s}) damages the performance of student networks, as validated in Section~\ref{Ablation}.

\paragraph{Mimicry loss.} On the one hand, we consider transferring hierarchical self-supervised augmented distributions generated from $L$ auxiliary classifiers of the teacher network to corresponding $L$ auxiliary classifiers of the student network, respectively. The transfer performs in a one-to-one manner by KL-divergence loss $D_{\rm{KL}}$. The loss is formulated as Eq.~(\ref{mini_ss}), where $\tau^{2}$ is used to retain the gradient contributions unchanged~\cite{hinton2015distilling}.
\begin{equation}
\mathcal{L}_{kl\_q}=\frac{1}{M}\sum_{j=1}^{M}\sum_{l=1}^{L}\tau^{2}D_{\rm{KL}}(\bm{q}^{T}_{l}(t_{j}(\bm{x});\tau)\parallel \bm{q}^{S}_{l}(t_{j}(\bm{x});\tau))
\label{mini_ss}
\end{equation}
Benefiting from Eq.~(\ref{mini_ss}), one can expect that the student network gains comprehensive guidances by unified self-supervised knowledge and the original class full-supervised knowledge. The informative knowledge is derived from multi-scale intermediate feature maps encapsulated in hidden layers of the high-capacity teacher network. 

On the other hand, we transfer the original class probability distributions generated from the final layer between teacher and student. Specifically, we transfer the knowledge derived from both the normal and transformed data $\{t_{j}(\bm{x})\}_{j=1}^{M}$, where $t_{1}(\bm{x})=\bm{x}$. This loss is formulated as Eq.~(\ref{mini_s}). 
\begin{equation}
\mathcal{L}_{kl\_p}=\frac{1}{M}\sum_{j=1}^{M}\tau^{2}D_{\rm{KL}}(\bm{p}^{T}(t_{j}(\bm{x});\tau)\parallel \bm{p}^{S}(t_{j}(\bm{x});\tau))
\label{mini_s}
\end{equation}
We do not explicitly force the student backbone $f^{S}(\cdot)$ to fit the transformed data in task loss for preserving the normal classification capability. But mimicking the side product of predictive class probability distributions inferred from these transformed data from the teacher network is also beneficial for the self-supervised representational learning of student network, as validated in Section~\ref{Ablation}.

\paragraph{Overall loss.} We summarize the task loss and  mimicry loss as the overall loss $\mathcal{L}_{S}$ for training the student network.
\begin{equation}
\mathcal{L}_{S}=\mathbb{E}_{\bm{x}\in \mathcal{X}}[\mathcal{L}_{task}+\mathcal{L}_{kl\_q}+\mathcal{L}_{kl\_p}]
\label{overall}
\end{equation}
Following the wide practice, we set the hyper-parameter $\tau=1$ in task loss and $\tau=3$ in mimicry loss. Besides, we do not introduce other hyper-parameters.

\section{Experiments}
\subsection{Experimental Settings}
We conduct evaluations on standard CIFAR-100 and ImageNet~\cite{deng2009imagenet} benchmarks across the widely applied network families including ResNet~\cite{he2016deep}, WRN~\cite{zagoruyko2016wide}, VGG~\cite{simonyan2014very}, MobileNet~\cite{sandler2018mobilenetv2} and ShuffleNet~\cite{zhang2018shufflenet,ma2018shufflenet}. Some representative KD methods including KD~\cite{hinton2015distilling}, FitNet~\cite{romero2014fitnets}, AT~\cite{zagoruyko2016paying}, AB~\cite{heo2019knowledge}, VID~\cite{Sungsoo19Variational}, RKD~\cite{park2019relational}, SP~\cite{Tung2019Similarity}, CC~\cite{peng2019correlation}, CRD~\cite{tian2019contrastive} and SOTA SSKD~\cite{DBLP:conf/eccv/XuLLL20} are compared. For a fair comparison, all comparative methods are combined with conventional KD by default, and we adopt rotations $\{0^{\circ},90^{\circ},180^{\circ},270^{\circ}\}$ as the self-supervised auxiliary task as same as SSKD. We use the standard training settings following~\cite{DBLP:conf/eccv/XuLLL20} and report the mean result with a standard deviation over 3 runs. \textbf{\emph{The more detailed settings for reproducibility can be found in our released codes}}.

\subsection{Ablation Study}
\begin{figure}[tbp]  
	\centering  
	\includegraphics[width=1\linewidth]{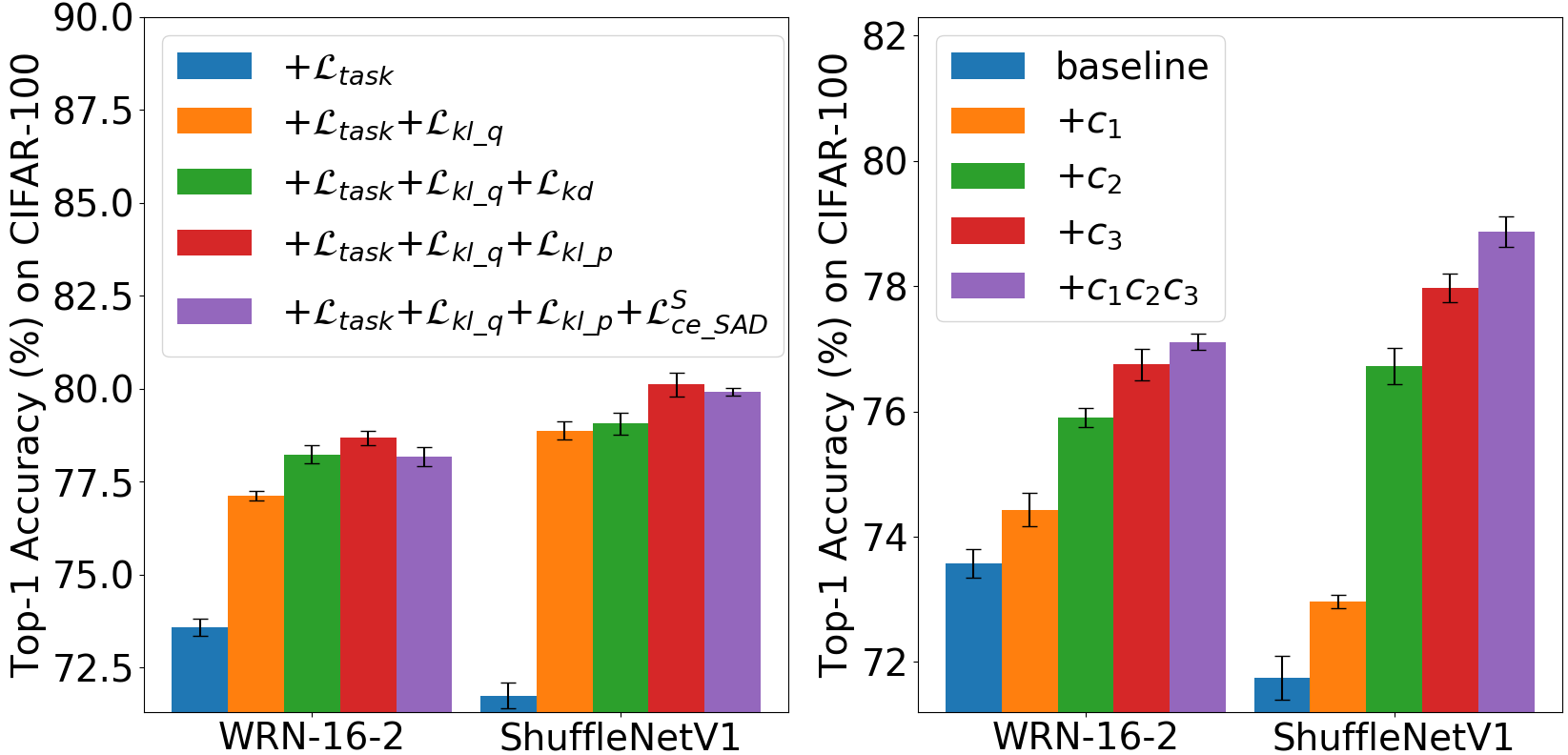}
	\caption{Ablation study of loss terms (\emph{\textbf{left}}) and auxiliary classifiers (\emph{\textbf{right}}) on the student networks WRN-16-2 and ShuffleNetV1 under the pre-trained teacher network WRN-40-2 on CIFAR-100.}  
	\label{ablation}
\end{figure}

\paragraph{Effect of loss terms.} As shown in Fig.~\ref{ablation} (left), applying hierarchical self-supervised augmented knowledge transfer through multiple auxiliary classifiers by loss $\mathcal{L}_{kl\_q}$ substantially boosts the accuracy upon the original task loss $\mathcal{L}_{task}$. We further compare $\mathcal{L}_{kl\_p}$  and $\mathcal{L}_{kd}$ upon $\mathcal{L}_{task}+\mathcal{L}_{kl\_q}$ to demonstrate the efficacy of transferring class probability distributions from additional transformed images. We find that $\mathcal{L}_{kl\_p}$ results in better accuracy gains than $\mathcal{L}_{kd}$, which suggests that transferring probabilistic class knowledge from those transformed images is also beneficial to feature representation learning. Finally, beyond the mimicry loss, we also explore whether the self-supervised augmented task loss $\mathcal{L}_{ce\_SAD}^{S}$ can be integrated into the overall task loss to train student networks. After adding $\mathcal{L}_{ce\_SAD}^{S}$ upon the above losses, the performance of the student network is dropped slightly. We speculate that mimicking soft self-supervised augmented distribution from the teacher by $\mathcal{L}_{kl\_q}$ is good enough to learn rich self-supervised knowledge. Extra learning from hard one-hot distribution by $\mathcal{L}_{ce\_SAD}^{S}$ may interfere with the process of self-supervised knowledge transfer.
\paragraph{Effect of auxiliary classifiers.}We append several auxiliary classifiers to the network with various depths to learn and transfer diverse self-supervised augmented distributions extracted from hierarchical features. To examine this practice, we first individually evaluate each auxiliary classifier. As shown in Fig.~\ref{ablation} (right), we can observe that each auxiliary classifier is beneficial to performance improvements. Moreover, the auxiliary classifier attached in the deeper layer often achieves more accuracy gains than that in the shallower layer, which can be attributed to more informative semantic knowledge encoded in high-level features. Finally, using all auxiliary classifiers can maximize accuracy gains.
\label{Ablation}
\begin{table*}
	\centering
	\resizebox{1\linewidth}{!}{
		\begin{tabular}{cc|c|ccc|ccccccc|cc}
			\toprule
			Teacher  & Student &Acc&Teacher  &Teacher* &Student &KD &AT & CC& SP& RKD&CRD &SSKD&Ours&Ours* \\
			\midrule
			\multirow{2}{*}{ResNet-34}  & \multirow{2}{*}{ResNet-18} &Top-1 &73.31& 75.48 &69.75&70.66 &70.70&69.96&70.62&71.34&71.38&\underline{71.62}&72.16&\textbf{72.39}\\
			&  &Top-5 &91.42 & 92.67 &89.07 &89.88&90.00&89.17&89.80&90.37&90.49&\underline{90.67}&90.85&\textbf{91.00} \\

			\bottomrule
	\end{tabular}}
	\caption{Top-1 accuracy (\%) comparison on ImageNet. The compared results are from~\protect\cite{DBLP:conf/eccv/XuLLL20}.}
	\label{imagenetx}
\end{table*}

\begin{table*}
	\centering
	\resizebox{0.91\linewidth}{!}{
		\begin{tabular}{l|ccccccccccc|c}
			\toprule
			Transferred Dataset & Baseline &KD &FitNet&AT &AB& VID& RKD&SP& CC&CRD &SSKD&Ours  \\
			\midrule
			CIFAR-100$\rightarrow $ STL-10 & 67.76  &67.90 &69.41&67.37&67.82&69.29&69.74&68.96&69.13&70.09&\underline{71.03}&\textbf{74.66} \\
			CIFAR-100$\rightarrow $ TinyImageNet & 34.69 &34.15&36.04&34.44&34.79&36.09&37.21&35.69&36.43&38.17&\underline{39.07}&\textbf{42.57} \\
			\bottomrule
	\end{tabular}}
	\caption{Linear classification accuracy (\%) of transfer learning on the student MobileNetV2 pre-trained using the teacher VGG-13. }
	\label{transfer}
\end{table*}

\begin{table}
	\centering
	\resizebox{0.65\linewidth}{!}{
		\begin{tabular}{cccccc}
			\toprule
			Baseline & KD& CRD &SSKD&Ours  \\
			\midrule
			76.18 &77.06  &77.36 &\underline{77.60}&\textbf{78.45}\\
			\bottomrule
	\end{tabular}}
	\caption{Comparison of detection mAP (\%) on Pascal VOC using ResNet-18 as the backbone pre-trained by various KD methods.}
	\label{detection}
\end{table}

\subsection{Comparison with State-Of-The-Arts}
\paragraph{Results on CIFAR-100 and ImageNet.} We compare our HSAKD with SOTA representative distillation methods across various teacher-student pairs with the same and different architectural styles on CIFAR-100 in Table~\ref{Comparisonx} and on ImageNet in Table~\ref{imagenetx}. Interestingly, using $\mathcal{L}_{ce\_SAD}^{T}$ as an auxiliary loss can improve the teacher accuracy of the final classification. Compared to the original 'Teacher', 'Teacher*' achieves an average gain of 4.09\% across five teacher networks on CIFAR-100 and a top-1 gain of 2.17\% for ResNet-34 on ImageNet. For downstream student networks, 'Teacher*' leads to an average improvement of 1.15\% on CIFAR-100 and an improvement of 0.23\% on ImageNet than 'Teacher'. These results indicate that our proposed loss $\mathcal{L}_{T}$ can improve the performance of a given network and produce a more suitable teacher for KD to learn a better student.

Moreover, our HSAKD significantly outperforms the best-competing method SSKD across all network pairs with an average accuracy gain of 2.56\% on CIFAR-100 and a top-1 gain of 0.77\% on ImageNet. Compared with other SOTA methods, the superiority of HSAKD can be attributed to hierarchical self-supervised augmented knowledge distillation by the assistance with well-designed auxiliary classifiers. 
\paragraph{Transferability of Learned Representations.} Beyond the accuracy on the object dataset, we also expect the student network can produce the generalized feature representations that transfer well to other unseen semantic recognition datasets. To this end, we freeze the feature extractor pre-trained on the upstream CIFAR-100, and then train two linear classifiers based on frozen pooled features for downstream STL-10 and TinyImageNet respectively, following the common linear classification protocal~\cite{tian2019contrastive}. As shown in Table~\ref{transfer}, we can observe that both SSKD and HSAKD achieve better accuracy than other comparative methods, demonstrating that using self-supervised auxiliary tasks for distillation is conducive to generating better feature representations. Moreover, HSAKD can significantly outperform the best-competing SSKD by 3.63\% on STL-10 and 3.50\% on TinyImageNet. The results verify that encoding the self-supervised auxiliary task as an augmented distribution in our HASKD  has better supervision quality than the contrastive relationship in SSKD for learning good features.

\paragraph{Transferability for Object Detection.} We further evaluate the student network ResNet-18 pre-trained with the teacher ResNet-34 on ImageNet as a backbone to carry out downstream object detection on Pascal VOC. We use Faster-RCNN~\cite{ren2016faster} framework and follow the standard data preprocessing and finetuning strategy. The comparison on detection performance is shown in Table~\ref{detection}.  Our method outperforms the original baseline by 2.27\% mAP and the best-competing SSKD by 0.85\% mAP. These results verify that our method can guide a network to learn better feature representations for semantic recognition tasks.
\begin{table}
	\centering
	\resizebox{1.\linewidth}{!}{
		\begin{tabular}{c|ccc|c}
			\toprule
			Percentage & KD &CRD &SSKD&Ours  \\
			\midrule
			25\% & 65.15$_{(\pm 0.23)}$ &65.80$_{(\pm 0.61)}$ &\underline{67.82}$_{(\pm 0.30)}$&\textbf{68.50}$_{(\pm 0.24)}$ \\
			50\% & 68.61$_{(\pm 0.22)}$ &69.91$_{(\pm 0.20)}$ &\underline{70.08}$_{(\pm 0.13)}$&\textbf{72.18}$_{(\pm 0.41)}$ \\
			75\% & 70.34$_{(\pm 0.09)}$ &\underline{70.98}$_{(\pm 0.43)}$ &70.47$_{(\pm 0.14)}$&\textbf{73.26}$_{(\pm 0.11)}$  \\
			\bottomrule
	\end{tabular}}
	\caption{Top-1 accuracy (\%) comparison on CIFAR-100 under few-shot scenario with various percentages of training samples. We use the ResNet56-ResNet20 as the teacher-student pair for evaluation.}
	\label{few_shot}
\end{table}
\paragraph{Efficacy under Few-shot Scenario.} We compare our method with conventional KD and SOTA CRD and SSKD under few-shot scenarios by retaining 25\%, 50\% and 75\% training samples. For a fair comparison, we use the same data split strategy for each few-shot setting, while maintaining the original test set. As shown in Table~\ref{few_shot}, our method can consistently surpass others by large margins under various few-shot settings. Moreover, it is noteworthy that by using only 25\% training samples, our method can achieve comparable accuracy with the baseline trained on the complete set. This is because our method can effectively learn general feature representations from limited data. In contrast, previous methods often focus on mimicking the inductive bias from intermediate feature maps or cross-sample relationships, which may overfit the limited set and generalize worse to the test set.

\section{Conclusion}
We propose a self-supervised augmented task for KD and further transfer such rich knowledge derived from hierarchical feature maps leveraging well-designed auxiliary classifiers. Our method achieves SOTA performance on the standard image classification benchmarks in the field of KD. It can guide the network to learn well-general feature representations for semantic recognition tasks. Moreover, it has no hyper-parameters to be tuned and is easy to implement.

\appendix

\bibliographystyle{named}
\bibliography{ijcai21}

\title{Appendix of Hierarchical Self-supervised Augmented Knowledge Distillation}
\appendix

\maketitle
\appendix

\section{Dataset and Training Settings}
\paragraph{CIFAR-100.} CIFAR-100 is composed of 50K training images and 10K test images in 100 classes. We use the standard data augmentation, \emph{i.e.} padding to $40\times 40$, random cropping to $32\times 32$ and random horizontal flipping. We train all networks by the SGD optimizer with a momentum of 0.9, a batch size of 64 and a weight decay of $5\times 10^{-4}$. The initial learning rate starts at 0.05 and is decayed by a factor of 10 at 150, 180 and 210 epochs within the total 240 epochs. For MobileNetV2, ShuffleNetV1 and ShuffleNetV2, we use a learning rate of 0.01. We run experiments on an NVIDIA RTX 3090 GPU.
\paragraph{ImageNet.} ImageNet is composed of 1.2 million training images and 50K validation images in 1000 classes. We use the standard data augmentation, \emph{i.e.} randomly cropping the image into the size ranging from $(0.08,1.0)$ of the original size and a random aspect ratio ranging from  $(3/4,4/3)$ of the original aspect ratio. The cropped patch is resized to $224\times 224$ and is randomly horizontal flipped with a probability of 0.5. We train all networks by the SGD optimizer with a momentum of 0.9, a batch size of 256 and a weight decay of $1\times 10^{-4}$. The initial learning rate starts at 0.1 and is decayed by a factor of 10 at 30, 60 and 90 epochs within the total 100 epochs. We run experiments on several parallel NVIDIA Tesla V100 GPUs.
\paragraph{STL-10 and TinyImageNet.} STL-10 is composed of 5K labeled training images and 8K test images in 10 classes. TinyImageNet is composed of 100K training images and 10k test images in 200 classes. For data augmentation, we randomly crop the image into the size ranging from $(0.08,1.0)$ of the original size and a random aspect ratio ranging from  $(3/4,4/3)$ of the original aspect ratio. The cropped patch is resized to $32\times 32$ and is randomly horizontal flipped with a probability of 0.5. We train the linear classifiers by the SGD optimizer with a momentum of 0.9, a batch size of 64 and a weight decay of $0$. The initial learning rate starts at 0.1 and is decayed by a factor of 10 at 30, 60 and 90 epochs within the total 100 epochs. We run experiments on an NVIDIA RTX 3090 GPU.

All experiments are conducted based on the Pytorch framework.

\section{Architectural Design of Auxiliary Classifiers}

As discussed in the main paper, we attach one auxiliary classifier after each convolutional stage. The auxiliary classifier is sequentially composed of several convolutional stages, global average pooling (GAP) and a fully-connected (FC) layer, where the design of convolutional stages follows the original architecture. In principle, the convolutional architecture in each auxiliary classifier is the same as the original backbone after the corresponding convolutional stage for architectural identity to ensure the fine-to-coarse feature transformation. Assumed that a classification network $f(\cdot)$ contains $L$ convolutional stages $1\sim L$, the $l$-th auxiliary classifier is composed of the $l+1\sim L$ stages, where $l\in [1,L-1]$. Specifically, the architecture of the final $L$-th auxiliary classifier is composed of the $L$-th original convolutional stage but without down-sampling for preserving the spatial information. Moreover, because we utilize a self-supervised augmented label with rotations $\{0^{\circ},90^{\circ},180^{\circ},270^{\circ}\}$, the class dimension of the FC layer in each auxiliary classifier is 4 times compared with that of the original classification FC layer.

We illustrate the overall architecture of various networks with auxiliary classifiers involved in in the main paper, including the family of WRN (Table~\ref{wrn402}, \ref{wrn401},  \ref{wrn162})~\cite{zagoruyko2016wide}, ResNet (Table~\ref{resnet56}, \ref{resnet20},  \ref{resnet32}, \ref{resnet8}, \ref{resnet50}, \ref{resnet50_imagenet})~\cite{he2016deep}, VGG (Table~\ref{vgg_13})~\cite{simonyan2014very}, MobileNet (Table~\ref{mbnv2})~\cite{sandler2018mobilenetv2} and ShuffleNet (Table~\ref{ShuffleNetV1}, \ref{ShuffleNetV2})~\cite{zhang2018shufflenet,ma2018shufflenet}. For better readability, the style of the illustration of architectural details is followed by the original paper.
\begin{table*}
	\centering
	\begin{tabular}{c|c|c|c|c|c}
		\hline
		Layer name & Output size & $f(\cdot)$& $c_{1}(\cdot)$& $c_{2}(\cdot)$& $c_{3}(\cdot)$\\ 	\hline
		conv1&  32$\times$32 &  $3\times3, 16$ &-&-&-\\  \hline
		conv2\_x &  32$\times$32              & 
		$\begin{bmatrix}
		3\times 3, 32\\ 
		3\times 3, 32
		\end{bmatrix}\times 6 $  &  -&  - &  -  \\  \hline
		conv3\_x &  16$\times$16           & 
		$\begin{bmatrix}
		3\times 3, 64\\ 
		3\times 3, 64
		\end{bmatrix}\times 6 $  &  $\begin{bmatrix}
		3\times 3, 64\\ 
		3\times 3, 64
		\end{bmatrix}\times 6 $ & - & -  \\  \hline
		conv4\_x &  8$\times$8           & 
		$\begin{bmatrix}
		3\times 3, 128\\ 
		3\times 3, 128
		\end{bmatrix}\times 6 $  &  $\begin{bmatrix}
		3\times 3, 128\\ 
		3\times 3, 128
		\end{bmatrix}\times 6 $&  $\begin{bmatrix}
		3\times 3, 128\\ 
		3\times 3, 128
		\end{bmatrix}\times 6 $& -  \\  \hline
		conv5\_x &  8$\times$8           & - &  -&  - &  $\begin{bmatrix}
		3\times 3, 128\\ 
		3\times 3, 128
		\end{bmatrix}\times 6 $  \\  \hline
		Classifier &  1$\times$1  &  GAP, 100-D FC&  GAP, 400-D FC &  GAP, 400-D FC&  GAP, 400-D FC  \\  \hline
	\end{tabular}
	\caption{Architectural details of WRN-40-2 with auxiliary classifiers for CIFAR-100 classification.}
	\label{wrn402}
\end{table*}

\begin{table*}
	\centering
	\begin{tabular}{c|c|c|c|c|c}
		\hline
		Layer name & Output size & $f(\cdot)$& $c_{1}(\cdot)$& $c_{2}(\cdot)$& $c_{3}(\cdot)$\\ 	\hline
		conv1&  32$\times$32 &  $3\times3, 16$ &-&-&-\\  \hline
		conv2\_x &  32$\times$32              & 
		$\begin{bmatrix}
		3\times 3, 16\\ 
		3\times 3, 16
		\end{bmatrix}\times 6 $  &  -&  - &  -  \\  \hline
		conv3\_x &  16$\times$16           & 
		$\begin{bmatrix}
		3\times 3, 32\\ 
		3\times 3, 32
		\end{bmatrix}\times 6 $  &  $\begin{bmatrix}
		3\times 3, 32\\ 
		3\times 3, 32
		\end{bmatrix}\times 6 $ & - & -  \\  \hline
		conv4\_x &  8$\times$8           & 
		$\begin{bmatrix}
		3\times 3, 64\\ 
		3\times 3, 64
		\end{bmatrix}\times 6 $  &  $\begin{bmatrix}
		3\times 3, 64\\ 
		3\times 3, 64
		\end{bmatrix}\times 6 $&  $\begin{bmatrix}
		3\times 3, 64\\ 
		3\times 3, 64
		\end{bmatrix}\times 6 $& -  \\  \hline
		conv5\_x &  8$\times$8           & - &  -&  - &  $\begin{bmatrix}
		3\times 3, 64\\ 
		3\times 3, 64
		\end{bmatrix}\times 6 $  \\  \hline
		Classifier &  1$\times$1  &  GAP, 100-D FC&  GAP, 400-D FC &  GAP, 400-D FC&  GAP, 400-D FC  \\  \hline
	\end{tabular}
	\caption{Architectural details of WRN-40-1  with auxiliary classifiers for CIFAR-100 classification.}
	\label{wrn401}
\end{table*}

\begin{table*}
	\centering
	\begin{tabular}{c|c|c|c|c|c}
		\hline
		Layer name & Output size & $f(\cdot)$& $c_{1}(\cdot)$& $c_{2}(\cdot)$& $c_{3}(\cdot)$\\ 	\hline
		conv1&  32$\times$32 &  $3\times3, 16$ &-&-&-\\  \hline
		conv2\_x &  32$\times$32              & 
		$\begin{bmatrix}
		3\times 3, 32\\ 
		3\times 3, 32
		\end{bmatrix}\times 2 $  &  -&  - &  -  \\  \hline
		conv3\_x &  16$\times$16           & 
		$\begin{bmatrix}
		3\times 3, 64\\ 
		3\times 3, 64
		\end{bmatrix}\times 2 $  &  $\begin{bmatrix}
		3\times 3, 64\\ 
		3\times 3, 64
		\end{bmatrix}\times 2 $ & - & -  \\  \hline
		conv4\_x &  8$\times$8           & 
		$\begin{bmatrix}
		3\times 3, 128\\ 
		3\times 3, 128
		\end{bmatrix}\times 2 $  &  $\begin{bmatrix}
		3\times 3, 128\\ 
		3\times 3, 128
		\end{bmatrix}\times 2 $&  $\begin{bmatrix}
		3\times 3, 128\\ 
		3\times 3, 128
		\end{bmatrix}\times 2 $& -  \\  \hline
		conv5\_x &  8$\times$8           & - &  -&  - &  $\begin{bmatrix}
		3\times 3, 128\\ 
		3\times 3, 128
		\end{bmatrix}\times 2 $  \\  \hline
		Classifier &  1$\times$1  &  GAP, 100-D FC&  GAP, 400-D FC &  GAP, 400-D FC&  GAP, 400-D FC  \\  \hline
	\end{tabular}
	\caption{Architectural details of WRN-16-2 with auxiliary classifiers for CIFAR-100 classification.}
	\label{wrn162}
\end{table*}

\begin{table*}
	\centering
	\begin{tabular}{c|c|c|c|c|c}
		\hline
		Layer name & Output size & $f(\cdot)$& $c_{1}(\cdot)$& $c_{2}(\cdot)$& $c_{3}(\cdot)$\\ 	\hline
		conv1&  32$\times$32 &  $3\times3, 16$ &-&-&-\\  \hline
		conv2\_x &  32$\times$32              & 
		$\begin{bmatrix}
		3\times 3, 16\\ 
		3\times 3, 16
		\end{bmatrix}\times 9 $  &  -&  - &  -  \\  \hline
		conv3\_x &  16$\times$16           & 
		$\begin{bmatrix}
		3\times 3, 32\\ 
		3\times 3, 32
		\end{bmatrix}\times 9 $  &  $\begin{bmatrix}
		3\times 3, 32\\ 
		3\times 3, 32
		\end{bmatrix}\times 9 $ & - & -  \\  \hline
		conv4\_x &  8$\times$8           & 
		$\begin{bmatrix}
		3\times 3, 64\\ 
		3\times 3, 64
		\end{bmatrix}\times 9 $  &  $\begin{bmatrix}
		3\times 3, 64\\ 
		3\times 3, 64
		\end{bmatrix}\times 9 $&  $\begin{bmatrix}
		3\times 3, 64\\ 
		3\times 3, 64
		\end{bmatrix}\times 9 $& -  \\  \hline
		conv5\_x &  8$\times$8           & - &  -&  - &  $\begin{bmatrix}
		3\times 3, 64\\ 
		3\times 3, 64
		\end{bmatrix}\times 9 $  \\  \hline
		Classifier &  1$\times$1  &  GAP, 100-D FC&  GAP, 400-D FC &  GAP, 400-D FC&  GAP, 400-D FC  \\  \hline
	\end{tabular}
	\caption{Architectural details of resnet56 with auxiliary classifiers for CIFAR-100 classification.}
	\label{resnet56}
\end{table*}

\begin{table*}
	\centering
	\begin{tabular}{c|c|c|c|c|c}
		\hline
		Layer name & Output size & $f(\cdot)$& $c_{1}(\cdot)$& $c_{2}(\cdot)$& $c_{3}(\cdot)$\\ 	\hline
		conv1&  32$\times$32 &  $3\times3, 16$ &-&-&-\\  \hline
		conv2\_x &  32$\times$32              & 
		$\begin{bmatrix}
		3\times 3, 16\\ 
		3\times 3, 16
		\end{bmatrix}\times 3 $  &  -&  - &  -  \\  \hline
		conv3\_x &  16$\times$16           & 
		$\begin{bmatrix}
		3\times 3, 32\\ 
		3\times 3, 32
		\end{bmatrix}\times 3 $  &  $\begin{bmatrix}
		3\times 3, 32\\ 
		3\times 3, 32
		\end{bmatrix}\times 3 $ & - & -  \\  \hline
		conv4\_x &  8$\times$8           & 
		$\begin{bmatrix}
		3\times 3, 64\\ 
		3\times 3, 64
		\end{bmatrix}\times 3 $  &  $\begin{bmatrix}
		3\times 3, 64\\ 
		3\times 3, 64
		\end{bmatrix}\times 3 $&  $\begin{bmatrix}
		3\times 3, 64\\ 
		3\times 3, 64
		\end{bmatrix}\times 3 $& -  \\  \hline
		conv5\_x &  8$\times$8           & - &  -&  - &  $\begin{bmatrix}
		3\times 3, 64\\ 
		3\times 3, 64
		\end{bmatrix}\times 3 $  \\  \hline
		Classifier &  1$\times$1  &  GAP, 100-D FC&  GAP, 400-D FC &  GAP, 400-D FC&  GAP, 400-D FC  \\  \hline
	\end{tabular}
	\caption{Architectural details of resnet20 with auxiliary classifiers for CIFAR-100 classification.}
	\label{resnet20}
\end{table*}

\begin{table*}
	\centering
	\begin{tabular}{c|c|c|c|c|c}
		\hline
		Layer name & Output size & $f(\cdot)$& $c_{1}(\cdot)$& $c_{2}(\cdot)$& $c_{3}(\cdot)$\\ 	\hline
		conv1&  32$\times$32 &  $3\times3, 32$ &-&-&-\\  \hline
		conv2\_x &  32$\times$32              & 
		$\begin{bmatrix}
		3\times 3, 64\\ 
		3\times 3, 64
		\end{bmatrix}\times 5 $  &  -&  - &  -  \\  \hline
		conv3\_x &  16$\times$16           & 
		$\begin{bmatrix}
		3\times 3, 128\\ 
		3\times 3, 128
		\end{bmatrix}\times 5 $  &  $\begin{bmatrix}
		3\times 3, 128\\ 
		3\times 3, 128
		\end{bmatrix}\times 5 $ & - & -  \\  \hline
		conv4\_x &  8$\times$8           & 
		$\begin{bmatrix}
		3\times 3, 256\\ 
		3\times 3, 256
		\end{bmatrix}\times 5 $  &  $\begin{bmatrix}
		3\times 3, 256\\ 
		3\times 3, 256
		\end{bmatrix}\times 5 $&  $\begin{bmatrix}
		3\times 3, 256\\ 
		3\times 3, 256
		\end{bmatrix}\times 5 $& -  \\  \hline
		conv5\_x &  8$\times$8           & - &  -&  - &  $\begin{bmatrix}
		3\times 3, 256\\ 
		3\times 3, 256
		\end{bmatrix}\times 5 $  \\  \hline
		Classifier &  1$\times$1  &  GAP, 100-D FC&  GAP, 400-D FC &  GAP, 400-D FC&  GAP, 400-D FC  \\  \hline
	\end{tabular}
	\caption{Architectural details of resnet32$\times$4 with auxiliary classifiers for CIFAR-100 classification.}
	\label{resnet32}
\end{table*}

\begin{table*}
	\centering
	\begin{tabular}{c|c|c|c|c|c}
		\hline
		Layer name & Output size & $f(\cdot)$& $c_{1}(\cdot)$& $c_{2}(\cdot)$& $c_{3}(\cdot)$\\ 	\hline
		conv1&  32$\times$32 &  $3\times3, 32$ &-&-&-\\  \hline
		conv2\_x &  32$\times$32              & 
		$\begin{bmatrix}
		3\times 3, 64\\ 
		3\times 3, 64
		\end{bmatrix}\times 1 $  &  -&  - &  -  \\  \hline
		conv3\_x &  16$\times$16           & 
		$\begin{bmatrix}
		3\times 3, 128\\ 
		3\times 3, 128
		\end{bmatrix}\times 1 $  &  $\begin{bmatrix}
		3\times 3, 128\\ 
		3\times 3, 128
		\end{bmatrix}\times 1 $ & - & -  \\  \hline
		conv4\_x &  8$\times$8           & 
		$\begin{bmatrix}
		3\times 3, 256\\ 
		3\times 3, 256
		\end{bmatrix}\times 1 $  &  $\begin{bmatrix}
		3\times 3, 256\\ 
		3\times 3, 256
		\end{bmatrix}\times 1 $&  $\begin{bmatrix}
		3\times 3, 256\\ 
		3\times 3, 256
		\end{bmatrix}\times 1 $& -  \\  \hline
		conv5\_x &  8$\times$8           & - &  -&  - &  $\begin{bmatrix}
		3\times 3, 256\\ 
		3\times 3, 256
		\end{bmatrix}\times 1 $  \\  \hline
		Classifier &  1$\times$1  &  GAP, 100-D FC&  GAP, 400-D FC &  GAP, 400-D FC&  GAP, 400-D FC  \\  \hline
	\end{tabular}
	\caption{Architectural details of resnet8$\times$4 with auxiliary classifiers for CIFAR-100 classification.}
	\label{resnet8}
\end{table*}

\begin{table*}
	\centering
	\resizebox{1\linewidth}{!}{
		\begin{tabular}{c|c|c|c|c|c|c}
			\hline
			Layer name & Output size & $f(\cdot)$& $c_{1}(\cdot)$& $c_{2}(\cdot)$& $c_{3}(\cdot)$& $c_{4}(\cdot)$ \\ 	\hline
			conv1&  32$\times$32 &  $3\times3, 64$ &-&-&-&  - \\  \hline
			conv2\_x &  32$\times$32              & 
			$\begin{bmatrix}
			1\times 1, 64\\
			3\times 3, 64\\ 
			1\times 1, 256
			\end{bmatrix}\times 3 $  &  -&  - &  -&  -   \\  \hline
			conv3\_x &  16$\times$16           & 
			$\begin{bmatrix}
			1\times 1, 128\\
			3\times 3, 128\\ 
			1\times 1, 512
			\end{bmatrix}\times 4 $  &  $\begin{bmatrix}
			1\times 1, 128\\
			3\times 3, 128\\ 
			1\times 1, 512
			\end{bmatrix}\times 4 $ & - & - &  - \\  \hline
			
			conv4\_x &  8$\times$8           & 
			$\begin{bmatrix}
			1\times 1, 256\\
			3\times 3, 256\\ 
			1\times 1, 1024
			\end{bmatrix}\times 6 $  &  $\begin{bmatrix}
			1\times 1, 256\\
			3\times 3, 256\\ 
			1\times 1, 1024
			\end{bmatrix}\times 6 $&  $\begin{bmatrix}
			1\times 1, 256\\
			3\times 3, 256\\ 
			1\times 1, 1024
			\end{bmatrix}\times 6 $& - &  -  \\  \hline
			
			conv5\_x &  4$\times$4          & 
			$\begin{bmatrix}
			1\times 1, 512\\
			3\times 3, 512\\ 
			1\times 1, 2048
			\end{bmatrix}\times 3$  & $\begin{bmatrix}
			1\times 1, 512\\
			3\times 3, 512\\ 
			1\times 1, 2048
			\end{bmatrix}\times 3$&  $\begin{bmatrix}
			1\times 1, 512\\
			3\times 3, 512\\ 
			1\times 1, 2048
			\end{bmatrix}\times 3 $&$\begin{bmatrix}
			1\times 1, 512\\
			3\times 3, 512\\ 
			1\times 1, 2048
			\end{bmatrix}\times 3$ & -  \\  \hline

			conv6\_x &  4$\times$4           & - &  -&  - &  - &$\begin{bmatrix}
			1\times 1, 512\\
			3\times 3, 512\\ 
			1\times 1, 2048
			\end{bmatrix}\times 3 $  \\  \hline
			Classifier &  1$\times$1  &  GAP, 100-D FC&  GAP, 400-D FC &  GAP, 400-D FC&  GAP, 400-D FC&  GAP, 400-D FC  \\  \hline
	\end{tabular}}
	\caption{Architectural details of ResNet-50 with auxiliary classifiers for CIFAR-100 classification.}
	\label{resnet50}
\end{table*}

\begin{table*}
	\centering
	\resizebox{1\linewidth}{!}{
		\begin{tabular}{c|c|c|c|c|c|c}
			\hline
			Layer name & Output size & $f(\cdot)$& $c_{1}(\cdot)$& $c_{2}(\cdot)$& $c_{3}(\cdot)$& $c_{4}(\cdot)$\\ 	\hline
			conv1\_x&  32$\times$32 &  $\begin{bmatrix}
			3\times 3\ \rm{conv}, 64\\ 
			3\times 3\ \rm{conv}, 64\\

		\end{bmatrix} $ &-&-&-\\  \hline
		
		conv2\_x &  16$\times$16              & 
		$\begin{bmatrix}
		2\times 2\ \rm{max pool}, \rm{stride}\ 2\\
		3\times 3\ \rm{conv}, 128\\ 
		3\times 3\ \rm{conv}, 128
		
	\end{bmatrix} $  & $\begin{bmatrix}
		2\times 2\ \rm{max pool}, \rm{stride}\ 2\\
		3\times 3\ \rm{conv}, 128\\ 
		3\times 3\ \rm{conv}, 128
		
	\end{bmatrix} $&  - &  -  \\  \hline
	
	conv3\_x &  8$\times$8           & 
	$\begin{bmatrix}
	2\times 2\ \rm{max pool}, \rm{stride}\ 2\\
	3\times 3\ \rm{conv}, 256\\ 
	3\times 3\ \rm{conv}, 256

\end{bmatrix} $  &  $\begin{bmatrix}
	2\times 2\ \rm{max pool}, \rm{stride}\ 2\\
	3\times 3\ \rm{conv}, 256\\ 
	3\times 3\ \rm{conv}, 256

\end{bmatrix} $& $\begin{bmatrix}
	2\times 2\ \rm{max pool}, \rm{stride}\ 2\\
	3\times 3\ \rm{conv}, 256\\ 
	3\times 3\ \rm{conv}, 256

\end{bmatrix} $& -  \\  \hline

conv4\_x &  4$\times$4          & 
$\begin{bmatrix}
2\times 2\ \rm{max pool}, \rm{stride}\ 2\\
3\times 3\ \rm{conv}, 512\\ 
3\times 3\ \rm{conv}, 512\\
3\times 3\ \rm{conv}, 512\\ 
3\times 3\ \rm{conv}, 512

\end{bmatrix} $  &  $\begin{bmatrix}
2\times 2\ \rm{max pool}, \rm{stride}\ 2\\
3\times 3\ \rm{conv}, 512\\ 
3\times 3\ \rm{conv}, 512\\
3\times 3\ \rm{conv}, 512\\ 
3\times 3\ \rm{conv}, 512

\end{bmatrix} $ &  $\begin{bmatrix}
2\times 2\ \rm{max pool}, \rm{stride}\ 2\\
3\times 3\ \rm{conv}, 512\\ 
3\times 3\ \rm{conv}, 512\\
3\times 3\ \rm{conv}, 512\\ 
3\times 3\ \rm{conv}, 512

\end{bmatrix} $ &  $\begin{bmatrix}
2\times 2\ \rm{max pool}, \rm{stride}\ 2\\
3\times 3\ \rm{conv}, 512\\ 
3\times 3\ \rm{conv}, 512\\
3\times 3\ \rm{conv}, 512\\ 
3\times 3\ \rm{conv}, 512

\end{bmatrix} $ &- \\  \hline

conv6\_x &  4$\times$4           & - &  -&  - &  -&$\begin{bmatrix}
3\times 3\ \rm{conv}, 512\\ 
3\times 3\ \rm{conv}, 512\\
3\times 3\ \rm{conv}, 512\\ 
3\times 3\ \rm{conv}, 512
\end{bmatrix} $  \\  \hline
Classifier &  1$\times$1  &  GAP, 100-D FC&  GAP, 400-D FC &  GAP, 400-D FC&  GAP, 400-D FC&  GAP, 400-D FC  \\  \hline
\end{tabular}}
\caption{Architectural details of VGG-13 with auxiliary classifiers for CIFAR-100 classification.}
\label{vgg_13}
\end{table*}

\begin{table*}
\centering
\resizebox{1\linewidth}{!}{
\begin{tabular}{ccc|cccc|cccc|cccc|cccc|cccc|}
\toprule
&&& \multicolumn{4}{c|}{$f(\cdot)$}& \multicolumn{4}{c|}{$c_{1}(\cdot)$}& \multicolumn{4}{c|}{$c_{2}(\cdot)$}& \multicolumn{4}{c|}{$c_{3}(\cdot)$}& \multicolumn{4}{c|}{$c_{4}(\cdot)$}\\ 	\midrule
Layer name & Input&Operator &$t$& $c$&$n$& $s$&$t$& $c$&$n$& $s$ &$t$& $c$&$n$& $s$&$t$& $c$&$n$& $s$&$t$& $c$&$n$& $s$\\ 	\midrule
conv1 & 32$^{2}\times$3&conv2d &-& 16&1& 1&-&-&-&-&-&-&-&-&-&-&-&-&-&-&-&-\\ 
\midrule
\multirow{2}{*}{conv2\_x} & 32$^{2}\times$8&bottleneck &1& 8&1& 1&-&-&-&-&-&-&-&-&-&-&-&-&-&-&-&-\\ 
& 32$^{2}\times$12&bottleneck &6& 12&2& 1&-&-&-&-&-&-&-&-&-&-&-&-&-&-&-&-\\ 
\midrule
conv3\_x & 32$^{2}\times$12&bottleneck &6& 16&3& 2&6& 16&3& 2&-&-&-&-&-&-&-&-&-&-&-&-\\ 
\midrule
\multirow{2}{*}{conv4\_x} & 16$^{2}\times$16&bottleneck &6& 32&4& 2&6& 32&4& 2&6& 32&4& 2&-&-&-&-&-&-&-&-\\ 
& 8$^{2}\times$32&bottleneck &6& 48&3& 1&6& 48&3& 1&6& 48&3& 1&-&-&-&-&-&-&-&-\\ 
\midrule

\multirow{2}{*}{conv5\_x}& 8$^{2}\times$48&bottleneck &6& 80&3& 2&6& 80&3& 2&6& 80&3& 2&6& 80&3& 2&-&-&-&-\\ 	

& 4$^{2}\times$80&bottleneck &6& 160&1& 1&6& 160&1& 1&6& 160&1& 1&6& 160&1& 1&-&-&-&-\\ 
\midrule
\multirow{2}{*}{conv6\_x}& 4$^{2}\times$160&bottleneck &-&-&-&-&-&-&-&-&-&-&-&-&-&-&-&-&6& 80&3& 1\\ 	

& 4$^{2}\times$80&bottleneck &-&-&-&-&-&-&-&-&-&-&-&-&-&-&-&-&6& 160&1& 1\\ 
\midrule
conv7 & 4$^{2}\times$160&conv2d 1x1 &-& 640&1& 1&-& 640&1& 1&-& 640&1& 1&-& 640&1& 1&-& 640&1& 1\\ 
\midrule
\multirow{2}{*}{classifier} & 4$^{2}\times$640&avgpool 7x7 &-&-&1& -&-&-&1& -&-&-&1& -&-&-&1& -&-&-&1& -\\ 
& 1$^{2}\times$640&conv2d 1x1 &-&100&1& -&-&400&1& -&-&400&1& -&-&400&1& -&-&400&1& -\\

\bottomrule

\end{tabular}}
\caption{Architectural details of MobileNetV2 with auxiliary classifiers for CIFAR-100 classification.}
\label{mbnv2}
\end{table*}

\begin{table*}
\centering
\resizebox{1\linewidth}{!}{
\begin{tabular}{cc|ccc|ccc|ccc|ccc|cccc}
\toprule
&&\multicolumn{3}{|c|}{$f(\cdot)$}&\multicolumn{3}{|c|}{$c_{1}(\cdot)$}&\multicolumn{3}{|c|}{$c_{2}(\cdot)$}&\multicolumn{3}{|c|}{$c_{3}(\cdot)$}&$f(\cdot)$&$c_{1}(\cdot)$&$c_{2}(\cdot)$&$c_{3}(\cdot)$ \\
Layer&Output size&KSize&Stride&Repeat&KSize&Stride&Repeat&KSize&Stride&Repeat&KSize&Stride&Repeat&\multicolumn{4}{|c|}{Output channels}\\
\midrule
Image&32$\times$32 &&&&&&&&&&&&&\multicolumn{4}{|c}{3}\\
Conv1&32$\times$32 &3$\times$3&1&1&&&&&&&&&&\multicolumn{4}{|c}{24}\\
\midrule
Stage2&16$\times$16 &&2&1&&&&&&&&&&\multicolumn{4}{|c}{240}\\
&16$\times$16 &&1&3&&&&&&&&&&\multicolumn{4}{|c}{240}\\
\midrule
Stage3&8$\times$8 &&2&1&&2&1&&&&&&&\multicolumn{4}{|c}{480}\\
&8$\times$8 &&1&7&&1&7&&&&&&&\multicolumn{4}{|c}{480}\\
\midrule
Stage4&4$\times$4 &&2&1&&2&1&&2&1&&&&\multicolumn{4}{|c}{960}\\
&4$\times$4 &&1&3&&1&3&&1&3&&&&\multicolumn{4}{|c}{960}\\
\midrule
Stage5&4$\times$4 &&&&&&&&&&&1&1&\multicolumn{4}{|c}{960}\\
&4$\times$4 &&&&&&&&&&&1&3&\multicolumn{4}{|c}{960}\\
\midrule
GlobalPool&1$\times$1&7$\times$7&&&7$\times$7&&&7$\times$7&&&7$\times$7&&\\
FC&&&&&&&&&&&&&&100&400&400&400\\
\bottomrule
\end{tabular}}
\caption{Architectural details of ShuffleNetV1 with auxiliary classifiers for CIFAR-100 classification, where the group $g=3$.}
\label{ShuffleNetV1}
\end{table*}

\begin{table*}
\centering
\resizebox{1\linewidth}{!}{
\begin{tabular}{cc|ccc|ccc|ccc|ccc|cccc}
\toprule
&&\multicolumn{3}{|c|}{$f(\cdot)$}&\multicolumn{3}{|c|}{$c_{1}(\cdot)$}&\multicolumn{3}{|c|}{$c_{2}(\cdot)$}&\multicolumn{3}{|c|}{$c_{3}(\cdot)$}&$f(\cdot)$&$c_{1}(\cdot)$&$c_{2}(\cdot)$&$c_{3}(\cdot)$ \\
Layer&Output size&KSize&Stride&Repeat&KSize&Stride&Repeat&KSize&Stride&Repeat&KSize&Stride&Repeat&\multicolumn{4}{|c|}{Output channels}\\
\midrule
Image&32$\times$32 &&&&&&&&&&&&&\multicolumn{4}{|c}{3}\\
Conv1&32$\times$32 &3$\times$3&1&1&&&&&&&&&&\multicolumn{4}{|c}{24}\\
\midrule
Stage2&16$\times$16 &&2&1&&&&&&&&&&\multicolumn{4}{|c}{116}\\
&16$\times$16 &&1&3&&&&&&&&&&\multicolumn{4}{|c}{116}\\
\midrule
Stage3&8$\times$8 &&2&1&&2&1&&&&&&&\multicolumn{4}{|c}{232}\\
&8$\times$8 &&1&7&&1&7&&&&&&&\multicolumn{4}{|c}{232}\\
\midrule
Stage4&4$\times$4 &&2&1&&2&1&&2&1&&&&\multicolumn{4}{|c}{464}\\
&4$\times$4 &&1&3&&1&3&&1&3&&&&\multicolumn{4}{|c}{464}\\
\midrule
Stage5&4$\times$4 &&&&&&&&&&&1&1&\multicolumn{4}{|c}{464}\\
&4$\times$4 &&&&&&&&&&&1&3&\multicolumn{4}{|c}{464}\\
\midrule
Conv6&4$\times$4 &1$\times$1&1&1&1$\times$1&1&1&1$\times$1&1&1&1$\times$1&1&1&\multicolumn{4}{|c}{1024}\\
\midrule
GlobalPool&1$\times$1&7$\times$7&&&7$\times$7&&&7$\times$7&&&7$\times$7&&\\
FC&&&&&&&&&&&&&&100&400&400&400\\
\bottomrule
\end{tabular}}
\caption{Architectural details of ShuffleNetV2 with auxiliary classifiers for CIFAR-100 classification.}
\label{ShuffleNetV2}
\end{table*}

\begin{table*}
	\centering
	\resizebox{1\linewidth}{!}{
		\begin{tabular}{c|c|c|c|c|c|c}
			\hline
			Layer name & Output size & $f(\cdot)$& $c_{1}(\cdot)$& $c_{2}(\cdot)$& $c_{3}(\cdot)$& $c_{4}(\cdot)$\\ 	\hline
			conv1&  112$\times$112 &  $7\times7, 64$, stride 2 &-&-&-\\  \hline
			\multirow{2}{*}{conv2\_x} &  \multirow{2}{*}{56$\times$56}              & 
			3$\times$3, max pool, stride 2  &  -&  - &  - &  - \\  \cline{3-7}
			&&$\begin{bmatrix}
			3\times 3, 64\\ 
			3\times 3, 64
			\end{bmatrix}\times 2 $&  -&  - &  -&  -\\  \hline

			conv3\_x &  28$\times$28           & 
			$\begin{bmatrix}
			3\times 3, 128\\ 
			3\times 3, 128
			\end{bmatrix}\times 2 $  &  $\begin{bmatrix}
			3\times 3, 128\\ 
			3\times 3, 128
			\end{bmatrix}\times 2 $ & - & - &  - \\  \hline
			
			conv4\_x &  14$\times$14           & 
			$\begin{bmatrix}
			3\times 3, 256\\ 
			3\times 3, 256
			\end{bmatrix}\times 2 $  &  $\begin{bmatrix}
			3\times 3, 256\\ 
			3\times 3, 256
			\end{bmatrix}\times 2 $&  $\begin{bmatrix}
			3\times 3, 256\\ 
			3\times 3, 256
			\end{bmatrix}\times 2 $& - &  - \\  \hline
			
			conv5\_x &  7$\times$7         & 
			$\begin{bmatrix}
			3\times 3, 512\\ 
			3\times 3, 512
			\end{bmatrix}\times 2$  & $\begin{bmatrix}
			3\times 3, 512\\ 
			3\times 3, 512
			\end{bmatrix}\times 2$&  $\begin{bmatrix}
			3\times 3, 512\\ 
			3\times 3, 512
			\end{bmatrix}\times 2 $& $\begin{bmatrix}
			3\times 3, 512\\ 
			3\times 3, 512
			\end{bmatrix}\times 2 $ &  - \\  \hline

			conv6\_x &  7$\times$7          & - &  -&  - & - & $\begin{bmatrix}
			3\times 3, 512\\ 
			3\times 3, 512
			\end{bmatrix}\times 2 $  \\  \hline
			Classifier &  1$\times$1  &  GAP, 1000-D FC&  GAP, 4000-D FC &  GAP, 4000-D FC&  GAP, 4000-D FC &  GAP, 4000-D FC \\  \hline
	\end{tabular}}
	\caption{Architectural details of ResNet-18 with auxiliary classifiers for ImageNet classification.}
	\label{resnet18_imagenet}
\end{table*}

\begin{table*}
	\centering
	\resizebox{1\linewidth}{!}{
		\begin{tabular}{c|c|c|c|c|c|c}
			\hline
			Layer name & Output size & $f(\cdot)$& $c_{1}(\cdot)$& $c_{2}(\cdot)$& $c_{3}(\cdot)$& $c_{4}(\cdot)$\\ 	\hline
			conv1&  112$\times$112 &  $7\times7, 64$, stride 2 &-&-&-\\  \hline
			\multirow{2}{*}{conv2\_x} &  \multirow{2}{*}{56$\times$56}              & 
			3$\times$3, max pool, stride 2  &  -&  - &  - &  - \\  \cline{3-7}
			&&$\begin{bmatrix}
			3\times 3, 64\\ 
			3\times 3, 64
			\end{bmatrix}\times 3 $&  -&  - &  -&  -\\  \hline

			conv3\_x &  28$\times$28           & 
			$\begin{bmatrix}
			3\times 3, 128\\ 
			3\times 3, 128
			\end{bmatrix}\times 4 $  &  $\begin{bmatrix}
			3\times 3, 128\\ 
			3\times 3, 128
			\end{bmatrix}\times 4 $ & - & - &  - \\  \hline
			
			conv4\_x &  14$\times$14           & 
			$\begin{bmatrix}
			3\times 3, 256\\ 
			3\times 3, 256
			\end{bmatrix}\times 6 $  &  $\begin{bmatrix}
			3\times 3, 256\\ 
			3\times 3, 256
			\end{bmatrix}\times 6 $&  $\begin{bmatrix}
			3\times 3, 256\\ 
			3\times 3, 256
			\end{bmatrix}\times 6 $& - &  - \\  \hline
			
			conv5\_x &  7$\times$7         & 
			$\begin{bmatrix}
			3\times 3, 512\\ 
			3\times 3, 512
			\end{bmatrix}\times 3$  & $\begin{bmatrix}
			3\times 3, 512\\ 
			3\times 3, 512
			\end{bmatrix}\times 3$&  $\begin{bmatrix}
			3\times 3, 512\\ 
			3\times 3, 512
			\end{bmatrix}\times 3$& $\begin{bmatrix}
			3\times 3, 512\\ 
			3\times 3, 512
			\end{bmatrix}\times 3$ &  - \\  \hline

			conv6\_x &  7$\times$7          & - &  -&  - & - & $\begin{bmatrix}
			3\times 3, 512\\ 
			3\times 3, 512
			\end{bmatrix}\times 3 $  \\  \hline
			Classifier &  1$\times$1  &  GAP, 1000-D FC&  GAP, 4000-D FC &  GAP, 4000-D FC&  GAP, 4000-D FC &  GAP, 4000-D FC \\  \hline
	\end{tabular}}
	\caption{Architectural details of ResNet-34 with auxiliary classifiers for ImageNet classification.}
	\label{resnet34_imagenet}
\end{table*}

\end{document}